\documentclass[11pt,a4paper]{article}
\usepackage[hyperref]{emnlp2020}
\usepackage{times}
\usepackage{latexsym}

\usepackage{multirow} 

\usepackage{graphicx}

\usepackage{color}
\usepackage{amsfonts,amsmath,amssymb,amsthm}
\usepackage{bm,nicefrac}

\definecolor{new_red}{RGB}{200,0,0}
\definecolor{yellow_dark}{RGB}{200,120,0}
\definecolor{new_green}{RGB}{40,160,0}
\definecolor{new_blue}{RGB}{0,70,180}
\definecolor{new_violet}{RGB}{120,0,120}
\definecolor{new_gray}{RGB}{110,100,100}

\usepackage{microtype}

\aclfinalcopy 
\def\aclpaperid{1205} 

\title{Top-Rank-Focused Adaptive Vote Collection for the Evaluation of Domain-Specific Semantic Models}

\author{Pierangelo Lombardo$^{1}$, Alessio Boiardi$^{1}$, Luca Colombo$^1$, \\ {\bf Angelo Schiavone$^1$, and Nicolò Tamagnone}$^1$ \\
  $^1$ Intervieweb S.r.l. (Zucchetti group), Turin, Italy \\
   \texttt{\{pierangelo.lombardo, alessio.boiardi, luca.colombo,}\\ 
   \texttt{angelo.schiavone, nicolo.tamagnone\}@intervieweb.it}
}

\date{}

\begin{document}
\maketitle
\begin{abstract}
The growth of domain-specific applications of semantic models, boosted by the recent achievements of unsupervised embedding learning algorithms, demands domain-specific evaluation datasets.
In many cases, content-based recommenders being a prime example, these models are required to rank words or texts according to their semantic relatedness to a given concept, with particular focus on top ranks.
In this work, we give a threefold contribution to address these requirements: ({\it i}) we define a protocol for the construction, based on adaptive pairwise comparisons, of a relatedness-based evaluation dataset tailored on the available resources and optimized to be particularly accurate in top-rank evaluation;
({\it ii}) we define appropriate metrics, extensions of well-known ranking correlation coefficients, to evaluate a semantic model via the aforementioned dataset by taking into account the greater significance of top ranks.
Finally, ({\it iii}) we define a stochastic transitivity model to simulate semantic-driven pairwise comparisons, which confirms the effectiveness of the proposed dataset construction protocol.
\end{abstract}

\section{Introduction}

In recent years, we have been witnessing a growth of Natural Language Processing (NLP) applications in a wide range of specific domains, such as recruiting \cite{inda,qin2018enhancing}, law \cite{sugathadasa2017synergistic}, oil and gas \cite{nooralahzadeh2018evaluation}, social media analysis \cite{alrashdi2019deep}, online education \cite{dessi2019evaluating}, and biomedical \cite{patel2020t109}.
Embedding-based models have been playing a crucial role in this   specialization, as they allow the application of the same learning algorithm to a variety of different corpora of unlabeled texts, obtaining domain-specific models \cite{bengio2003neural,bojanowski2017enriching,devlin2018bert,mikolov2013efficient,mikolov2017advances,mikolov2013distributed,pennington2014glove}.

The evaluation and validation of a domain-specialized model requires  manually-annotated domain-specific datasets \cite{bakarov2018survey,lastra2019reproducible,taieb2019survey}.
However, the construction of such datasets is a very resource-consuming process, and particular care is needed to ensure their ability to evaluate the desired features \cite{bakarov2018survey,taieb2019survey,wang2019evaluating}.
In particular, it is fundamental to carefully consider the so-called {\it downstream task} (i.e., the final purpose of the model), because the appropriate evaluation metric depends on this task \cite{bakarov2018survey,blanco2013repeatable,halpin2010evaluating,rogers2018s,wang2019evaluating}.

Semantic similarity and relatedness are related but distinct notions in linguistics, the first being associated with concepts  which share taxonomic properties and being maximized by synonyms; 
on the other hand, semantically related concepts can share any kind of semantic relation, including antonym \cite{cai2010effective,harispe2015semantic}.
These  notions underlie the downstream tasks of countless NLP applications, including information retrieval \cite{akmal2014ontology,chen2017semantic,gurevych2007electronic,hliaoutakis2006information,ji2017using,lopez2017interpretable,srihari2000intelligent,uddin2013semantic}, content-based recommendation \cite{de2008integrating,de2015semantics,lops2011content}, semantic matching \cite{giunchiglia2004s,li2014semantic,wan2016deep},
ontology learning and knowledge management \cite{aouicha2016derivation,georgiev2018enhancing,jiang2014semantic,sanchez2008learning}, and word sense disambiguation \cite{aouicha2016wsd,patwardhan2003using}.

In view of the widespread of these applications, we propose a methodology to construct appropriate domain-specific datasets and metrics to assess the accuracy of relatedness and similarity estimations.
In particular, due to its suitability for non-expert human annotation, we mainly focus on semantic relatedness; however, the proposed protocol can be easily extended to semantic similarity.

A standard approach to evaluate a relatedness-based model is the comparison of the semantic ranking it produces with the
corresponding ranking determined from human annotations.
However, 
the relevance of rank mismatches may depend on the involved positions; in particular, top ranks are considered more important in many contexts, two prominent examples being content-based recommenders \cite{de2008integrating,de2015semantics,lops2011content,mladenic1999text} and semantic matching  \cite{giunchiglia2004s,li2014semantic,wan2016deep}. The greater significance of top ranks compared with low ranks is actually a pretty common phenomenon, as it can be argued from the attempts to overweight the former in the context of ranking correlation
\cite{blest2000theory,pinto2005weighted,dancelli2013two,iman1987measure,maturi2008new,shieh1998weighted,vigna2015weighted,webber2010similarity}. 

Our contribution is framed within the requirement to create domain-specific datasets to evaluate semantic relatedness measure with particular focus on top ranks and is threefold. 
({\it i}) In Section \ref{sec:dataset}, we define a protocol for the construction, based on adaptive pairwise comparisons, of a relatedness-based evaluation dataset tailored on the available resources and optimized to be particularly accurate in top-rank evaluation.
({\it ii}) In Section \ref{sec:weighted_ranks}, we define appropriate metrics to evaluate a semantic model via the aforementioned dataset by taking into account the greater significance of top ranks; the proposed metrics are extensions of well-known ranking correlation measures and they can be used to compare rankings, independently from their origin, whenever top ranks are particularly important.
Finally, ({\it iii}) in Section \ref{sec:vote_simul}, we define a stochastic model to simulate semantic-driven pairwise comparisons, 
whose predictions (described in Section \ref{sec:results}) confirm
 the effectiveness of the proposed dataset construction protocol; more in detail, we adapt a stochastic transitivity model, originally defined in the context of comparative judgment, in order to make it suitable for either similarity-driven or relatedness-driven comparisons.

\section{Dataset Construction}\label{sec:dataset}

In this section, we describe and justify a methodology to construct a dataset for the evaluation of a domain-specific relatedness-based model.
Relatedness-based evaluation -- known as {\it intrinsic evaluation} in the context of embedding-based models --  requires the construction of a dataset of human annotations, which may be collected via two different approaches. 
The former relies on a small group of linguistic experts to create a gold standard dataset, which is reliable but very expensive and, due to the subjectivity of relatedness and to the limited number of annotations, highly susceptible to bias and lack of statistical significance \cite{blanco2013repeatable,faruqui2016problems}. The latter relies on a large group of non-experts, typically associated with a crowdsourcing service (e.g., Amazon MTurk, ProlificAcademic, SocialSci, CrowdFlower, ClickWorker, CrowdSource), it is typically more affordable, and it has been proven to be repeatable and reliable \cite{blanco2013repeatable}.

\begin{figure}[!ht]
\centering
\includegraphics[width=0.95\columnwidth]{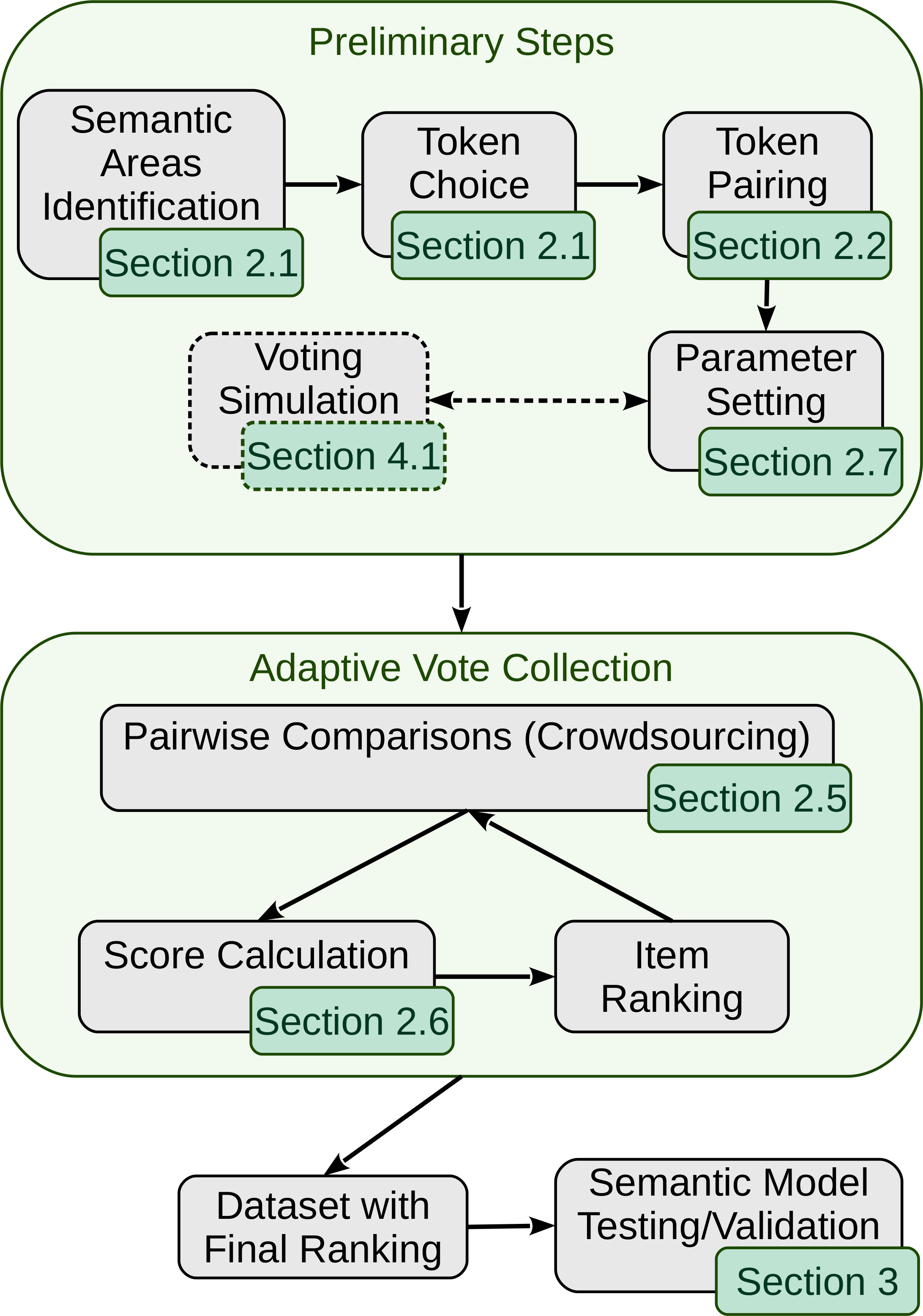}
\caption{Main steps for the practical construction of a dataset within the proposed approach.} \label{fig:flowchart}
\end{figure}

In the next sections we describe and justify a protocol to construct a dataset based on semantic relatedness between pairs of tokens%
\footnote{Hereafter, we refer as {\it token} to a word or a sequence of words that should be considered together, as they identify a single concept (e.g., {\it machine learning}).}
collected via a crowdsourcing approach.
To simplify the reading of the paper,
Figure~\ref{fig:flowchart} shows the main steps for the practical construction of a dataset within the proposed approach, while Table~\ref{tab:notation} reports a summary of the most frequently used symbols.

\begin{table}[!ht]
\centering
\caption{
Most commonly used symbols. 
}\label{tab:notation}
\begin{tabular}{l|l}
\multicolumn{2}{l}{Dataset Construction} \\
\hline
$N_{\rm tok}$ & Number of tokens\\
$N_{\rm items}$ & \begin{tabular}[c]{@{}l@{}}Number of items 
(i.e., pairs of tokens)\end{tabular} \\
$N_{\rm voters}$ & Number of voters\\
$N_{\rm comp}$ & Number of pairwise comparisons\\
$n_{\rm b}$ & Number of ballots\\
$\alpha$ & Fraction of selected items\\
$i,j$ & Indices specifying the item\\
$M$  & \begin{tabular}[c]{@{}l@{}}Number of times an item is presented\\ \ \
to the voters in each ballot\end{tabular} \\
$k$ & Index specifying the ballot\\
$N_{\rm items}^{(k)}$ & Number of items in ballot $k$\\
$x_i^{(k)}$  & \begin{tabular}[c]{@{}l@{}}Borda score (i.e., win ratio)\\ \ \
for item $i$ in ballot $k$ \end{tabular} \\
$y_i^{(k)}$ & Rescaled score for item $i$ in ballot $k$\\
\noalign{\vskip 0.3mm} 
\hline\cline{1-2} \cline{1-2} \noalign{\vskip 0.6mm} 
\multicolumn{2}{l}{Evaluation Metrics} \\
\hline
$a_i$ ($b_i$) &  \begin{tabular}[c]{@{}l@{}} Rank of item $i$ according to \\ \ \
the first (second) ranking \end{tabular}\\
$\rho_{\rm w}$ & Weighted version of Spearman's $\rho$\\
$\tau_{\rm w}$ & Weighted version of Kendall's $\tau$\\
$w_i^a$ ($w_i^b$) & Weight associated to $a_i$ ($b_i$)\\
$n_0$ & Offset in weight calculation\\
\noalign{\vskip 0.3mm} 
\hline\cline{1-2} \cline{1-2} \noalign{\vskip 0.6mm} 
\multicolumn{2}{l}{Stochastic Transitivity Model} \\
\hline
$z_i$ & Underlying similarity of item $i$\\
$v$ & Index specifying the voter\\
$o_i^{(v)}$ & Opinion of voter $v$ about item $i$\\
$\eta_i^{(v)}$ & \begin{tabular}[c]{@{}l@{}} Gaussian-distributed random \\ \ \
variable characterizing voter $v$ \end{tabular}\\
$\sigma_v^*$ & Nonconformity level of voter $v$\\
$\epsilon_v$ & Probability of oversight for voter $v$
\end{tabular}
\end{table}

\subsection{Token Choice}\label{sec:choice_tokens}

The first step in the dataset construction is the choice of the tokens among which we want to estimate the semantic relatedness.
These tokens must be carefully chosen to 
represent the semantic areas typically involved in the downstream tasks \cite{bakarov2018survey,schnabel2015evaluation}. 
Moreover, it is well-known that models based on high-dimensional embeddings tend to incorrectly identify as a semantic nearest neighbor
to almost any concept one of a few
common tokens called {\it hubs} \cite{dinu2014improving,feldbauer2018fast,francois2007concentration,radovanovic2010hubs,radovanovic2010existence}. In order to detect this undesirable feature, which goes under the name of {\it hubness problem}, an evaluation dataset must contain a relevant amount of rare%
\footnote{A token can be considered {\it rare} within a particular domain if its frequency in a corpus of domain-specific texts is, e.g., lower than $10\%$ of the average token frequency in the corpus.}
tokens \cite{bakarov2018survey,blanco2013repeatable}.
Henceforth, we consider as a concrete example a content-based recommender system in the recruitment domain \cite{inda}; in this case, the designated tokens can be chosen among  hard/soft skills, job titles, and other tokens found in resumes and job descriptions, including a relevant fraction of rare tokens.

Another potential issue of 
using relatedness to evaluate semantic models, is associated to 
lexical ambiguity, i.e., to the lack of one-to-one correspondence between tokens and meanings~\cite{bakarov2018survey,faruqui2016problems,wang2019evaluating}. 
To mitigate this problem, we suggest 
identifying
a number of relevant semantic areas within the domain of interest and subdivide the tokens accordingly. For instance, {\it Sales \& Marketing}, {\it Computer-related}, {\it Workforce}, and {\it Work \& Welfare} are examples of semantic areas within the recruiting domain.

\subsection{Token Pairing}\label{sec:choice_pairs}

The random sampling of pairs in the whole vocabulary is known to produce a large amount of unrelated pairs \cite{taieb2019survey}, in contrast with the desired focus on the most related pairs.
A standard approach to overcome this problem is pair selection based on either known semantic relations or 
the frequency of tokens' co-occurrence within a corpus of texts. While the former information may be {\it a priori} unknown within the domain of interest, the latter may produce a bias in favor of distributional methods that compute relatedness based on similar 
knowledge sources
\cite{taieb2019survey}.

We suggest, therefore, separate token pairing in each of the semantic areas identified as described in Section~\ref{sec:choice_tokens}: 
in this case, the relatedness distribution is substantially shifted towards larger values
compared with random sampling in the whole vocabulary.
This shift is shown in Figure~\ref{fig:theoretical_scores} -- based on a word embedding created with the word2vec algorithm \cite{mikolov2013efficient,mikolov2013distributed} trained on a corpus of resumes --
where the relatedness distribution of pairs of distinct tokens selected within the same semantic area%
\footnote{More in detail, 990 pairs of distinct tokens (associated with 45 tokens) have been considered within the semantic area {\it Sales \& Marketing}.}
(red plus) is compared with that of pairs randomly generated in the whole vocabulary (purple diamonds).
The generation of all pairs of distinct tokens produces $N_{\rm tok} (N_{\rm tok} - 1) / 2$ pairs per semantic area,  $N_{\rm tok}$ being the number of tokens. Although the pairs can be reduced via random sampling, an accurate evaluation requires a large number of pairs.

\subsection{Vote Collection}

Once we have defined the pairs of tokens, which will be referred to as {\it items} hereafter, we want to rank them, with particular emphasis on top ranks, according to the opinions of a large number $N_{\rm voters}$ of non experts.
Due to the large number of items involved, the complete ranking of all items would be an unfeasible task for a human and it is convenient to reformulate it in terms of pairwise comparisons~\cite{furnkranz2010preference,heckel2019active,heckel2018approximate,jamieson2011active,negahban2017rank,park2015preference,wauthier2013efficient}.
Moreover, the complete exploration of the $N_{\rm items} (N_{\rm items} - 1) / 2$ pairs of distinct items
would be extremely expensive in terms of votes;
luckily enough, it has been proven to be non-necessary in many studies \cite{jamieson2011active,negahban2017rank,park2015preference,wauthier2013efficient}.

\citet{louviere1991best} (see also \citet{kiritchenko2017best}) proposed a faster alternative to pairwise comparisons, known as {\it best-worst scaling}.
In this case $n$-tuples (typically, $n=4$), rather than pairs, are presented to the voter, who is required to identify the best and the worst items in each tuple, according to the relatedness of the corresponding tokens. 
This approach's drawbacks
are a reduction, for $n>3$, in the control on which pairs are actually checked and an increase in the complexity of each vote, which is particularly unwanted in crowdsourcing vote collections.
For this reason, we rely on the standard pairwise comparison:  we  generate $N_{\rm comp}$ pairs of items 
(as described in Sections \ref{sec:uniform_vote} and \ref{sec:adaptive_vote}), each one to be presented to one voter, who is requested to identify the item formed by the most similar tokens.

\subsection{Uniform Item Selection}\label{sec:uniform_vote}

In our setting, each  item $i$ is presented to the voters a total number of times $M_i$, and we define a score 
\begin{equation}\label{eq:scores_uniform}
 x_i = \frac{n_i}{M_i},    
\end{equation}
where $n_i$ represents the total number of times item $i$ was the winner%
\footnote{Ties can be accounted for by defining $n_i=n_i^{\rm w} + n_i^{\rm t}/2$, where $n_i^{\rm w}$ ($n_i^{\rm t}$) represents the number of wins (ties) for item $i$.}
in the vote collection; note that $x_i$ corresponds to an empirical approximation of the average probability%
\footnote{The average is intended  over all voters.}
 -- known as {\it Borda score} in the context of social choice theory -- that item $i$ beats a randomly chosen item $j\neq i$, where the accuracy of the approximation increases as $M_i$ increases \cite{borda1784memoire,heckel2019active}.

In the absence of 
{\it a priori} knowledge on the expected scores, a reasonable approach for the data collection consists of presenting each item the same number $M_i$ of times to the voters. 
In this scenario, we randomly generate $N_{\rm comp}$ pairs of items, with the constraint that 
$M_i = 2 N_{\rm comp} / N_{\rm items}\,\forall i$.
The only way to increase $M_i$ -- which is a proxy of the accuracy of the $x_i$ score defined in Equation~\ref{eq:scores_uniform} -- is, therefore, to  increase the total number of comparisons $N_{\rm comp}$.

\subsection{Adaptive Item Selection}\label{sec:adaptive_vote}

\citet{crompvoets2019adaptive}, \citet{heckel2019active}, \citet{heckel2018approximate}, \citet{jamieson2011active}, and \citet{negahban2017rank} proposed so-called {\it adaptive approaches} to increase of the efficiency of pairwise comparisons by identifying, before each comparison,
the optimal pair of items to be compared based on the votes already collected, on the task to be solved (typically, finding the global ranking or a ranking-induced partition), and on assumptions on the vote distribution.

The application of an adaptive approach in our context requires two additional ingredients which, to the best of our knowledge, are still missing in the literature: ({\it i})
in order 
to avoid overfitting on the opinion of the fastest voters and
to allow simultaneous voting, the choice of the pairs to be presented must occur
in a few events, as each of these events causes a discontinuity in the vote collection
(namely, this choice requires to suspend the vote collection when the numbers of comparisons reach the desired distribution among the voters)%
; ({\it ii}) the goal is a selective increase in the precision (proxied by $M_i$) of top ranks (with no need of {\it a priori} knowledge on the semantic relatedness of the tokens), rather than a general improvement in the global ranking.
In Section~\ref{sec:weighted_ranks} we define an appropriate metric to quantify top-rank accuracy.

The key idea is to subdivide the voting procedure in $n_{\rm b}$ subsequent ballots in which pairwise comparisons, based on a list of pairs determined before the beginning of the ballot, are presented to the voters.
During the first ballot, the pairs are randomly drawn from all items, with the constraint that each item appears $M$ times%
\footnote{We recommend an even value for $M$; otherwise, if both $M$ and $N_{\rm items}$ are odd, one item should appear $M+1$ times.}%
, while in each subsequent ballot $k$, the pairs are drawn from the $N_{\rm items}^{(k)}$ top-rank items selected according to the results of the previous ballots, with analogous constraint. 
More in detail, we define
\begin{equation}\label{eq:n_items_k}
    N_{\rm items}^{(k)} = {\rm round}( \alpha N_{\rm items}^{(k-1)} ) \sim \alpha^{k-1} N_{\rm items},
\end{equation}
where $\alpha$ represents the fraction of items selected at each ballot.
Since each item contained in ballot $k$ appears $M$ times within the pairs of such ballot, the total number of comparisons can be written as
\begin{equation}\label{eq:n_votes}
    N_{\rm comp} = \sum_{k=1}^{n_{\rm b}}\frac{M N_{\rm items}^{(k)}}{2} \sim \frac{1-\alpha^{n_{\rm b}}}{2(1-\alpha)}MN_{\rm items}.
\end{equation}

Thus, each item $i$ which survives up to the last ballot,
is presented to the voters a number of times
\begin{equation}\label{eq:Mtop}
    M_{\rm top} = n_{\rm b} M 
    \sim (1-\alpha) n_{\rm b} M_{\rm unif},
\end{equation}
where $M_{\rm unif} = 2 N_{\rm comp} / N_{\rm items}$ 
is the number of comparisons per item in the case of uniform item selection with the same total number of comparisons; the last approximation holds whenever  $\alpha^{n_{\rm b}} \ll 1$, i.e., when the fraction of items which survive up to the last ballot is small. 
According to Equation~\ref{eq:Mtop}, 
the score precision for top-rank items can be increased by decreasing the fraction $\alpha$ of selected items or by increasing the number $n_{\rm b}$ of ballots;
in Section~\ref{sec:parameters} we discuss bounds on these values.

\subsection{Score Calculation}

At each ballot $k$ and for each item $i$ contained in $k$, we can evaluate a score $x_i^{(k)}$ defined as in Equation~\ref{eq:scores_uniform}.
Nonetheless, since the pool of competing items is narrowed around top ranks at each ballot, the winning chances of a given item $i$ decrease accordingly.
For this reason,  the expected value of $x_i^{(k)}$ is smaller than the expected value of $x_i^{(k-1)}$, and this discrepancy must be taken into account in order to average scores from different ballots.

We define therefore a rescaled score $y_i^{(k)} = f_{\rm resc}^{(k)}(x_i^{(k)})$, where $f_{\rm resc}$ is the identity function for $k=1$, while it is obtained as a linear interpolation between $\{ x_j^{(k)} \}$ and  $\{ \bar y_j^{(k-1)} \}$ for $k>1$, where $\bar y_i^{(k)}$ is defined as the average of all rescaled scores up to ballot $k$, i.e., 
\begin{equation}\label{eq:average_score_k}
    \bar y_i^{(k)} = \frac{1}{k} \sum_{k'=1}^k y_i^{(k')}.
\end{equation}
We enforce the $f_{\rm resc}^{(k)}(1) = 1$ constraint in the linear interpolation, obtaining%
\footnote{Since the winning chances of a given item $j$ decrease at each ballot, on average, $x_j^{(k)} \leq \bar y_j^{(k-1)}$. Heuristically, we expect that, on average, $[1-x_j^{(k)}] [1-\bar y_j^{(k-1)}] \leq [1 - x_j^{(k)}]^2$, which implies 
$\hat b^{(k)} \leq 1$, which in turns ensures $y_i^{(k)}\leq 1$.}%
, for $k>1$,
\begin{equation}\label{eq:score_rescaling}
    \left\{
    \begin{array}{ll}
      y_i^{(k)} &= 1 - \hat b^{(k)} + \hat b^{(k)} x_i^{(k)}\\
      \hat b^{(k)} &= \frac{\sum_j [1 - x_j^{(k)}][1 - \bar y_j^{(k-1)}]}{\sum_j[1 - x_j^{(k)}]^2} 
    \end{array}
    \right. \ .
\end{equation}

\begin{figure}
\centering
\includegraphics[width=\columnwidth]{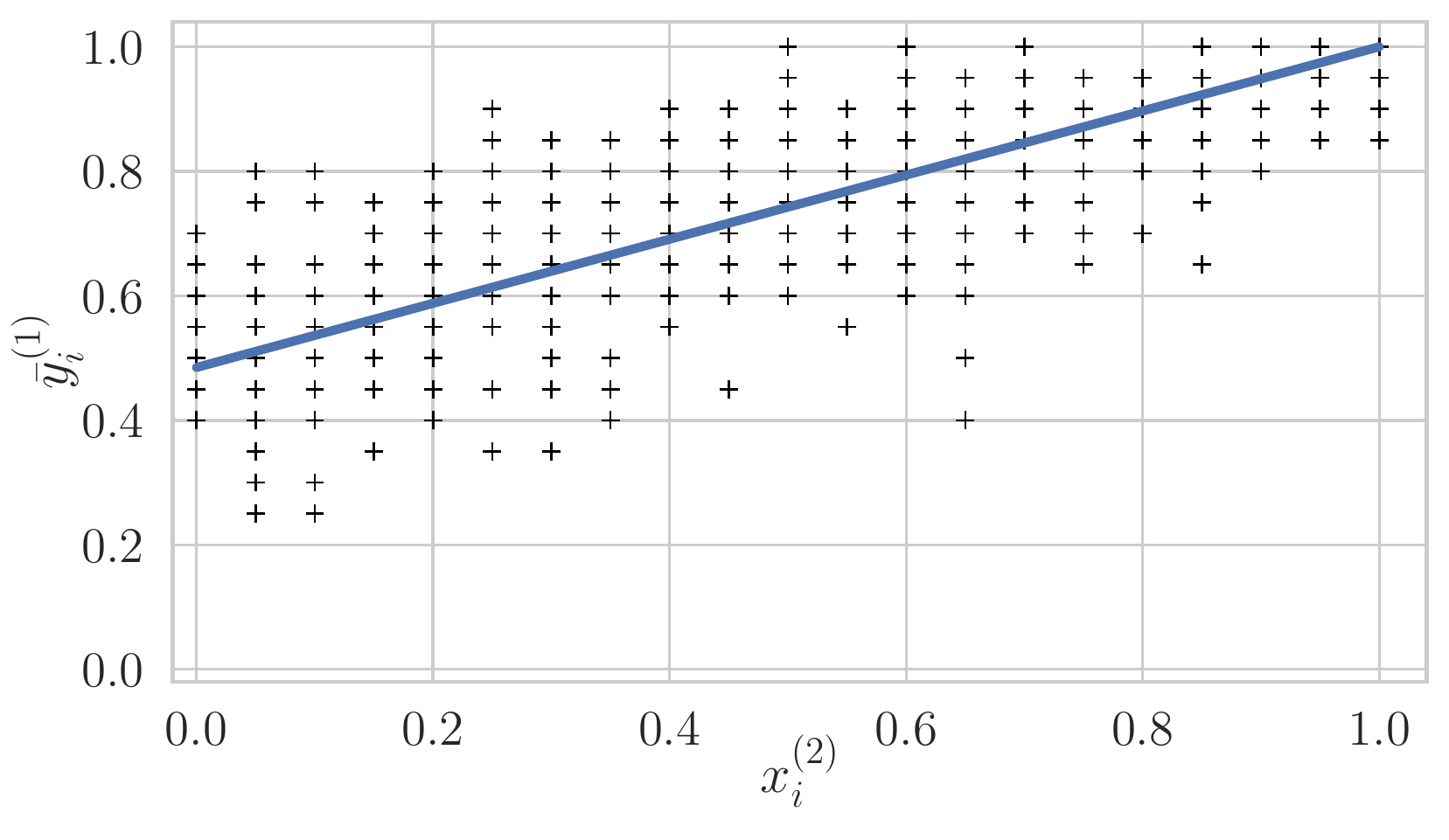}
\caption{The coordinates of this scatterplot represent the scores $x_i^{(2)}$ and $\bar y_i^{(1)} = x_i^{(1)}$, for each item $i$ present in the second ballot; the data are obtained with the model described in Section~\ref{sec:vote_simul} with exponential underlying similarity, $N_{\rm items}=990$, $\alpha=0.5$, $M=20$, $N_{\rm voters}=100$, $\sigma_v^*=0.1$, $\epsilon_v=0.01$.
The blue line represents the interpolation described in the text.} \label{fig:rescale}
\end{figure}

Figure~\ref{fig:rescale} shows an example of the interpolation on data simulated via the stochastic model described in Section~\ref{sec:vote_simul}.
Note that Equation~\ref{eq:average_score_k} provides a sequence of approximations of the Borda scores with accuracy increasing with $k$;
top ranks are expected to survive up to the last ballot, and therefore to be highly accurate.

\subsection{Choice of Parameter Values}\label{sec:parameters}

We provide here heuristics to identify ranges of values for the parameters of the adaptive approach. 

{\bf Number $\bm{n_{\rm b}}$ of Ballots.}
We need $n_{\rm b}\geq 2$ for the adaptive approach to be meaningful, while, 
to limit the discontinuities in the vote collection, a reasonable upper bound is $n_{\rm b}\lesssim 10$.
    
{\bf {Fraction $\bm{\alpha}$ of Selected Items.}}
As the purpose of the adaptive approach is to focus votes on top rank items, a reasonable request is to have no more than $10\%$ of items surviving up to the last ballot, which gives an upper bound $\alpha\lesssim(0.1)^{1/(n_{\rm b}-1)}$. On the other hand, at least two items must be present in the last ballot; according to Equation~\ref{eq:n_items_k}, this implies a lower bound $\alpha \gtrsim (2/N_{\rm items})^{1/(n_{\rm b}-1)}$. 
    
{\bf Number $\bm{N_{\rm comp}}$ of Comparisons.}
To achieve the desired precision 
(namely, the statistical significance of the averages) on top-rank scores, a reasonable request is $M_{\rm top} \gtrsim 100$; using Equation~\ref{eq:n_items_k} in the $\alpha^{n_{\rm b}} \ll 1$ limit, this implies $N_{\rm comp} \gtrsim 50 N_{\rm items} /[(1-\alpha)n_{\rm b}]$.
On the other hand, the only upper bound on $N_{\rm comp}$ is the cost of the voting, which can be estimated from Equation~\ref{eq:cost}.
        
{\bf Number $\bm{M}$ of Comparisons per Ballot.}
$M$ can be computed from Equation~\ref{eq:n_votes} once the other parameters have been fixed.
Note the existence of two competing phenomena: ({\it i}) decreasing $M$ (i.e., increasing $\alpha$) we increase the fluctuations in the $x_i$ scores defined in Equation~\ref{eq:scores_uniform}; ({\it ii}) for given $x_i$ fluctuations, decreasing $\alpha$ we increase the probability of top ranks' premature loss due to stricter selection.
A rigorous derivation of the optimal values of $M$ and $\alpha$ as a trade-off between these phenomena is beyond the scope of this section, as the bounds discussed above provide heuristic ranges.

\section{Evaluation Metrics}\label{sec:weighted_ranks}

Although the approximations of the Borda scores described in Sections~\ref{sec:uniform_vote} and \ref{sec:adaptive_vote} can be thought as estimates of the semantic relatedness,  we rely on rankings rather than scores to avoid inconsistency issues that frequently emerge in score comparisons 
\cite{ammar2011ranking,negahban2017rank}. 

\citet{kendall1948rank} proposed the quite general form for a ranking correlation coefficient 

\begin{equation}\label{eq:unweighted_coeff}
    \Gamma = \frac{\sum_{i,j}a_{ij}b_{ij}}{\sqrt{\sum_{ij}a_{ij}^2 \sum_{ij}b_{ij}^2}},
\end{equation}
where $a_{ij}$ (resp., $b_{ij}$) is a matrix that depends on the first (second) ranking to be compared, with indices $i,j$ running over all items. This definition
%
contains Spearman's $\rho$ \cite{spearman1961proof} as a particular case with $a_{ij}=a_j-a_i$ and $b_{ij} = b_j-b_i$, while Kendall's $\tau$ \cite{kendall1938new,kruskal1958ordinal} is obtained with $a_{ij}={\rm sign}(a_j-a_i)$ and $b_{ij} = {\rm sign}(b_j-b_i)$, where $\{a_i\}$ and $\{ b_i\}$ are the rankings to be compared.

In order to take into account the larger importance of top ranks in our context, we define weighted versions of $\rho$ and $\tau$, with increasing weight at the increasing of the rank position.
Namely, we define ({\it i}) $a_{ij} = \sqrt{w_iw_j}\,(a_j-a_i)$,  $b_{ij} = \sqrt{w_iw_j}\,(b_j-b_i)$ and ({\it ii}) $a_{ij} = \sqrt{w_iw_j}\,{\rm sign}(a_j-a_i)$,  $b_{ij} = \sqrt{w_iw_j}\,{\rm sign}(b_j-b_i)$, where $w_i$ is the normalized weight associated to the $i$-th position in the rankings. These coefficients can be rewritten respectively as
\begin{equation}\label{eq:weighted_coeff}
\begin{array}{ll}
     \rho_{\rm w} &= \frac{\sum_i w_i(a_i-\bar a)(b_i-\bar b)}{\sigma_a \sigma_b}\\
     \tau_{\rm w} &= \frac{\sum_{ij} w_i w_j\,{\rm sign}(a_j-a_i) \,{\rm sign}(b_j-b_i)}{Z(\{w_i\})} 
\end{array}\ \ ,
\end{equation}
where $\bar a=\sum_i w_i a_i$, $\sigma_a^2 = \sum_i w_i(a_i^2-\bar a^2)$, $\bar b=\sum_i w_i b_i$, $\sigma_b^2 = \sum_i w_i(b_i^2-\bar b^2)$, while $Z(\{w_i\})$ is a normalization factor, which corresponds to $1 - \sum_i w_i^2$ in the absence of ties; these metrics have been emerging, albeit some notation differences, as extensions
of the $\rho$ and $\tau$ coefficients to take into account the larger importance of top ranks
\cite{pinto2005weighted,dancelli2013two,vigna2015weighted}.

Different weighting schemes have been proposed in the literature  \cite{dancelli2013two,vigna2015weighted}; here we adopt the additive scheme 
\begin{equation}\label{eq:weights}
    w_i=\frac{w_i^a+w_i^b}{\sum_j(w_j^a+w_j^b)},
\end{equation}
with $w_i^a=f(a_i)$ and $w_i^b=f(b_i)$, where $f(n)$ is a monotonically decreasing function, in view of its ability in discriminating different rankings even when they only differ by the exchange of a top rank and a low rank \cite{dancelli2013two}.

A common choice is $f(n) = 1/n$ \cite{dancelli2013two,vigna2015weighted}; however, in the large $N_{\rm items}$ limit, it causes the divergence of the denominator in Equation~\ref{eq:weights} and makes thus any $w_i$ negligible.
This phenomenon is responsible for the decreased sensitivity on top ranks, observed by \citet{dancelli2013two}, in case of long rankings.
For this reason, we prefer to use $f(n;n_0) = 1 / (n+n_0)^2$, where the offset $n_0$ has been introduced to control the weight fraction associated to the first rank in the large $N_{\rm items}$ limit, i.e., $R(n_0) = f(1;n_0) / \sum_{n=1}^{\infty} f(n;n_0)$, which can be expressed as $R(n_0) = 1 / [ (n_0+1)^2\,  \psi^{(1)}(n_0+1)]$, where $\psi^{(1)}(x)$ is the first derivative of the digamma function.
With this choice, both $\rho_{\rm w}$ and $\tau_{\rm w}$ defined in Equation~\ref{eq:weighted_coeff} represent a family of correlation coefficients, depending on the value of $n_0$, whose choice depends on the particular task (namely, on the relative importance of the first rank).
The value $n_0=0$  causes an extremely high sensitivity on the first rank ($R(0) \sim 0.61$), which may be excessive; 
hereafter, we focus therefore on the value $n_0=2$, which appears to be a reasonable trade-off ($R(2) \sim 0.28$) that allows focusing on the first rank while avoiding neglecting other ranks.

The metrics $\rho_{\rm w}$ and $\tau_{\rm w}$ are suitable to compare rankings, whenever top ranks are particularly important; in particular, they can be used to evaluate a semantic model using a dataset produced as described in this paper.

\section{Evaluation of the Data-Collection Framework}

The collection of human annotations to construct a domain-specific dataset is a resource-consuming process, even within the proposed optimized data collection approach, whose \emph{person-hours} cost can be estimated as
\begin{equation}\label{eq:cost}
     C \sim \bar t_{\rm comp} N_{\rm comp},
\end{equation}
where $\bar t_{\rm comp}$ is the average time needed for a single comparison. 
For this reason, in Section~\ref{sec:vote_simul}, we define a stochastic model for semantic pairwise comparisons, which can be used to simulate the voting before the collection of human annotations, e.g., for checking or tuning the parameters of the data collection approach.
This stochastic model
will be used in Section~\ref{sec:results} to compare the effectiveness of the adaptive and the uniform approaches, using the metrics defined in Section \ref{sec:weighted_ranks}.

\subsection{Semantic Pairwise Comparisons}\label{sec:vote_simul}

We want to model $N_{\rm voters}$ voters to whom are proposed $N_{\rm comp}$ pairwise comparisons and who are asked to identify the item containing the most semantically related tokens.
The model will be used to reconstruct an approximate ranking of the items.

For the sake of mathematical simplicity, we firstly focus on similarity-driven comparisons, where the similarity $z$ takes value in the symmetric interval $[-1,1]$, where $z=1$, $0$, and $-1$ correspond respectively to synonyms, unrelated tokens, and antonyms.
The model will 
eventually be adapted to semantic relatedness by using the fact that, since antonyms correspond to 
semantically related tokens~\cite{cai2010effective,harispe2015semantic}, the absolute value $|z|$ is a reasonable proxy for semantic relatedness.

 \subsubsection{Similarity-Driven Comparisons.}\label{sec:similarity_model}
 
\begin{figure}
\centering
\includegraphics[width=\columnwidth]{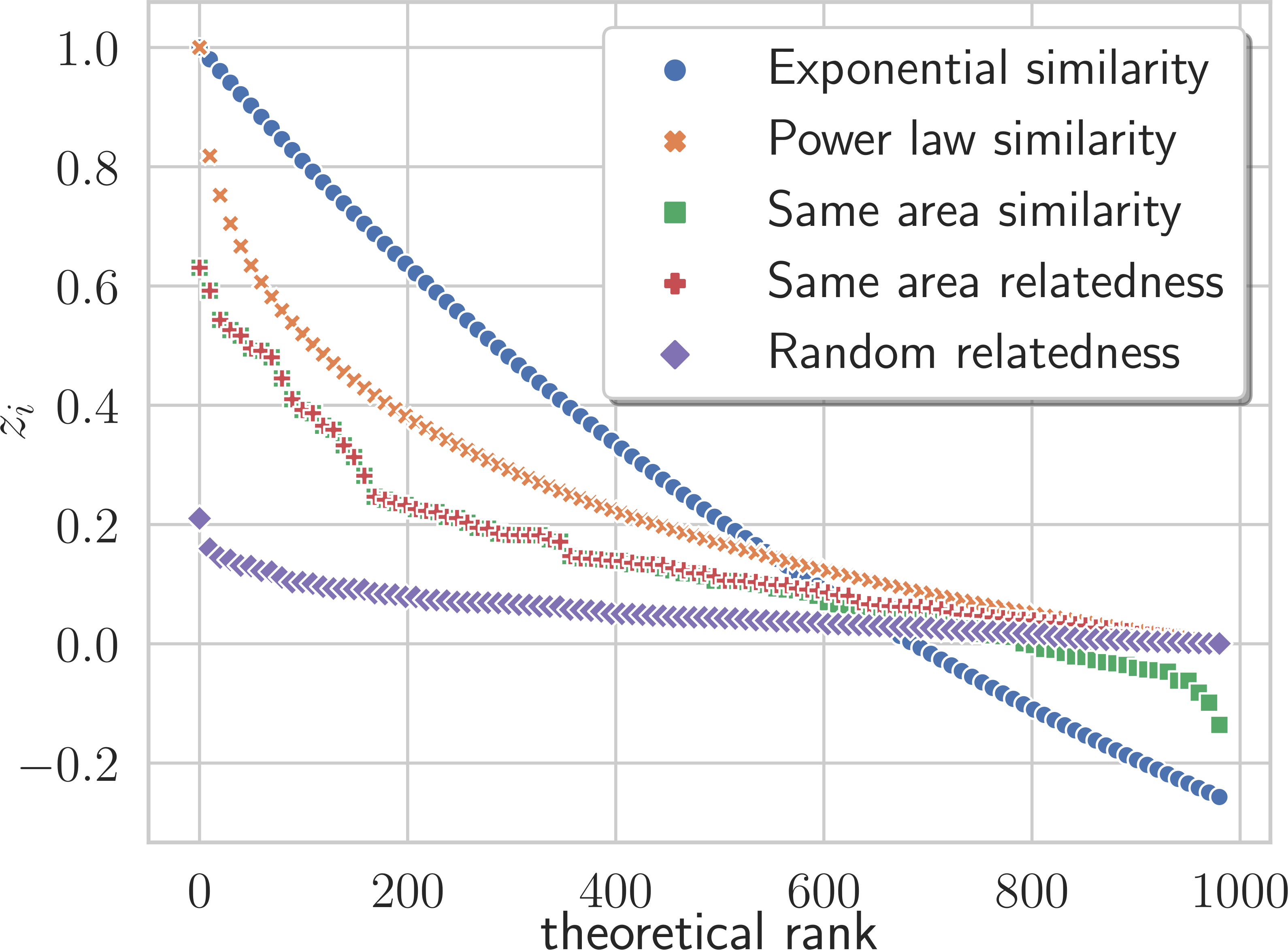}
\caption{We represent the three underlying similarity distributions described in Section \ref{sec:similarity_model} and the two relatedness distributions described in Section \ref{sec:choice_pairs}; relatedness is quantified by $|\cos\theta|$, where $\theta$ is the angle between the corresponding vectors in the embedding.}
\label{fig:theoretical_scores}
\end{figure}

A convenient way to model similarity-driven pairwise comparisons
assumes the existence of an underlying (unknown) similarity distribution $\{z_i\}$, which determines the theoretical ranks of the items, which in turn can be compared with the ranks estimated via the model.
We consider here three examples:({\it i}) an exponential  $z_i = 2\,{\rm exp}(-i/N_{\rm items})-1$, ({\it ii}) a power law $z_i = 2 / (1+\sqrt{i/N_{\rm items}})-1$, and ({\it iii}) the distribution of the cosine similarity%
\footnote{Despite a certain ambiguity observed by \citet{faruqui2016problems}, cosine similarity is typically considered a proxy of semantic similarity \cite{auguste2017evaluation,banjade2015lemon}.}
between pairs of tokens in the word embedding described in Section~\ref{sec:choice_pairs}; these distributions are represented in Figure~\ref{fig:theoretical_scores}.

A fundamental aspect to be considered in modelling similarity-driven pairwise comparisons is the 
task's subjectivity, as many potential linguistic, psychological, and social factors  could introduce biases \cite{bakarov2018survey,faruqui2016problems,gladkova2016intrinsic}.
A possible approach to account for this problem is via a stochastic transitivity model, firstly introduced in the context of comparative judgment of physical stimuli by \citet{thurstone1927law} (see also \citealp{cattelan2012models,ennis2016thurstonian});
this model describes the opinion $o_i^{(v)}$ of voter $v$ about item $i$ as a stochastic function $o_i^{(v)} = z_i + \sigma_v \eta_i^{(v)}$, where $z_i$ is the underlying similarity, $\eta_i^{(v)}$ is a Gaussian-distributed random variable with zero mean and unit variance, while $\sigma_v$ represents the {\it nonconformity amplitude}, i.e., the discrepancy between the voter's opinion and the underlying similarity.%
\footnote{Contrary to the original formulation, no covariance terms are present here, as the voters are supposed to be non-interacting.
Moreover, in the original formulation, voters and items are respectively referred to as {\it judges} and {\it stimuli}.
}

Here we define a modified version of the Thurstonian model, with a stochastic amplitude $\sigma_v$ depending on the underlying similarity, so that
\begin{equation}\label{eq:voter_similarity}
    o_i^{(v)} = F(z_i + \sigma_{v}(z_i)\,\eta_i^{(v)}), 
\end{equation}
where $F(x) = {\rm max}(-1,{\rm min}(1,x))$ has been introduced to enforce the constraint $-1\leq o_i^{(v)}\leq 1$, analogous to the one discussed above for $z$.
In the absence of $z_i$ dependence in the nonconformity amplitude, the probability $P_{\rm out}(z_i)$ to have $z_i+\sigma_v\eta_i$ outside the interval $[-1,1]$ would tend to $1/2$ as $z_i$ approaches one of the boundaries of the interval, causing, due to the $F$ constraint, the collapse of a relevant fraction of opinion $o_i$ to either $-1$ or $1$.
This degeneracy can be avoided with $\sigma_v(z_i)$ proportional to $1-z_i$  and $1+z_i$ as $z_i$ approaches $1$ and $-1$ respectively. Here we consider the simplest form with these features, i.e., $\sigma_{v}(z_i)=\sigma_v^*\, (1-z_i^2)$, which makes particularly sense in our context, where each item $i$ represents a pair of tokens, and the closest the similarity is to $z_i=1$ ($z_i=-1$), the higher is the relation (the opposition) between the tokens in the corresponding pair, and the stronger is expected to be the agreement in the voters' opinions on their similarity.

In order to increase the accuracy of the model, we introduce another source of randomness that represents the  {\it distraction level} of the voter, i.e., its tendency -- observed, e.g., by \citet{bakarov2018survey} and \citet{bruni2014multimodal} -- to unintentionally vote for the item perceived as lower rank. This tendency is accounted for by assuming that the result of a pairwise comparison presented to voter $v$ is actually the item with the highest perceived score $o_i^{(v)}$ with probability $1-\epsilon_v$ (with $\epsilon_v \ll 1$), while the other item is voted (oversight) with probability $\epsilon_v$.

The proposed model depends on the underlying similarity distribution%
\footnote{However, as shown in Table \ref{tab:results}, the dependence is mild.}%
, on the number of voters, on the random variables $\eta_i^{(v)}$, and on the voter-distinctive parameters $ \sigma_v^*$ and $\epsilon_v$, whose distribution could be experimentally determined by analyzing human voting. In the absence of such analysis, it seems reasonable to uniformly draw $ \sigma_v^*$ and $\epsilon_v$ from  ranges covering one order of magnitude to encompass human variability; heuristic upper bounds are $\epsilon_v\lesssim 0.05$ and $\sigma_v^*\lesssim 0.2$, as oversights are supposed to be rare, and the probability $P_{\rm out}(0)$ that two completely unrelated tokens ($z_i=0$) are deliberately considered as maximally related ($o_i^{(v)}=\pm 1$) should be extremely low
: the aforementioned bound corresponds indeed to $P_{\rm out}(0)\lesssim 0.001\%$.

\subsubsection{Relatedness-Driven Comparisons.}

As discussed in Section~\ref{sec:vote_simul}, we consider the absolute value of similarity as a proxy of relatedness.
The model defined in Section \ref{sec:similarity_model} is thus extended to relatedness-driven comparisons by
({\it i}) including an absolute value in Equation~\ref{eq:voter_similarity}, so that $ o_i^{(v)} = |F(z_i + \sigma_{v}^*(z_i)\,\eta_i^{(v)})|$ and 
({\it ii}) defining the theoretical rank of item $i$ according to $|z_i|$.

\begin{table*}[t]
\centering
\caption{Mean $\pm$ standard deviation (unbiased estimation over 50 simulations of relatedness-driven comparisons, as described in text) for 
$\rho_{\rm w}$ and $\tau_{\rm w}$ metrics, defined in Equation~\ref{eq:weighted_coeff}, and for Spearman's $\rho$ and Kendall's $\tau$ coefficients.}\label{tab:results}
\begin{tabular}{cccccc}
                                                   &           & $\bm{\rho_{\rm w}}$                         & $\bm{\tau_{\rm w}}$      & $\bm{\rho}$                & $\bm{\tau}$                \\ \hline \cline{1-6}
\multicolumn{1}{c}{\multirow{2}{*}{\ Exponential\ }} & Uniform   & $.778 \pm .058$                        & $\ -0.11 \pm .20 \ $ & $\ .8097 \pm .0088\ $ & $\ .6265 \pm .0091\ $ \\ 
\multicolumn{1}{c}{}                             & \ Adaptive\  & $\ .9452 \pm .0028\ $                  & $.66 \pm .17$     & $.8015 \pm  .0087$    & $.6330 \pm .0098$     \\ 
\hline
\multicolumn{1}{c}{\multirow{2}{*}{Power Law}}   & Uniform   & $.800 \pm .062$                        & $ -0.11 \pm .20 $     & $.9713 \pm .0013$     & $.8491 \pm .0035$     \\ 
\multicolumn{1}{c}{}                             & Adaptive  & $.9800 \pm .0014$ & $.63 \pm .18$       & $ .9632\pm .0019$     & $.8406 \pm .0040$     \\ 
\hline
\multicolumn{1}{c}{\multirow{2}{*}{Embedding}}   & Uniform   & $.741 \pm .058$                        & $ -0.11\pm .21$     & $.7229 \pm .0078$     & $.5463 \pm .0072$     \\ 
\multicolumn{1}{c}{}                             & Adaptive & $.9146 \pm .0042$                      & $.73 \pm .12$       & $.7258 \pm .0093$     & $.5611 \pm .0097$     \\ 
\hline \cline{1-6} \cline{1-6}
\end{tabular}
\end{table*}

\subsection{Results}\label{sec:results}

We estimated the accuracy of a data collection approach by comparing, via the metrics defined in Section \ref{sec:weighted_ranks}, the ranking that it produces with the underlying theoretical ranks.
We considered the semantic area described in Section~\ref{sec:choice_pairs}, containing 990 items, and we simulated a relatedness-driven data collection based on ({\it i}) the adaptive approach described in Section \ref{sec:adaptive_vote}, with $N_{\rm comp} = 39000$, $M = 20$, $\alpha = 0.5$, $n_{\rm b} = 7$ and ({\it ii}) the uniform approach described in Section \ref{sec:uniform_vote}, with the same total number of comparisons.
The voting was simulated with the stochastic model described in Section~\ref{sec:vote_simul}, with $N_{\rm voters} = 100$ and based on all three discussed distributions for the underlying similarity; for each voter $v$, the nonconformity level $\sigma_v^*$ and the distraction level $\epsilon_v$ were randomly chosen in the intervals $[0.02, 0.2]$ and $[0.005, 0.05]$ respectively, while each $\eta_i^{(v)}$ was randomly drawn from a normal distribution with zero mean and unit variance. Each simulation was repeated 50 times (by resampling all voters' parameters $\sigma_v^*$, $\epsilon_v$, and $\eta_i^{(v)}$ at each simulation) in order to obtain statistically significant results.

The code for our experiments is available at 
\url{https://github.com/intervieweb-datascience/adaptive-comp} and was run on a local machine equipped with an Intel Core i7-7700HQ (2.80GHz x8), with average runtimes of $30.9\,{\rm s}$ and $33.2\,{\rm s}$ respectively for the adaptive and the uniform approaches.
The results of the simulations are presented in Table~\ref{tab:results}, which contains, as measures of the accuracy of the proposed approaches, the $\rho_{\rm w}$ and $\tau_{\rm w}$ coefficients defined in Equation~\ref{eq:weighted_coeff} and discussed in Section~\ref{sec:weighted_ranks}; in order to check the overall rank accuracy, we also report the standard Spearman's $\rho$ and Kendall's $\tau$ coefficients.
For each coefficient, we report the average value and the unbiased estimator of the standard deviation over the 50 simulations.
The adaptive approach, compared with the uniform approach, determines a relevant increase in both $\rho_{\rm w}$ and $\tau_{\rm w}$ for any of the underlying similarity distributions considered, with no relevant changes
in the overall rank precision measured by $\rho$ and $\tau$.
Moreover, the results suggest that the proposed stochastic model is robust for changes in the underlying similarity distribution.

\begin{figure}[h!]
\centering
\includegraphics[width=\columnwidth]{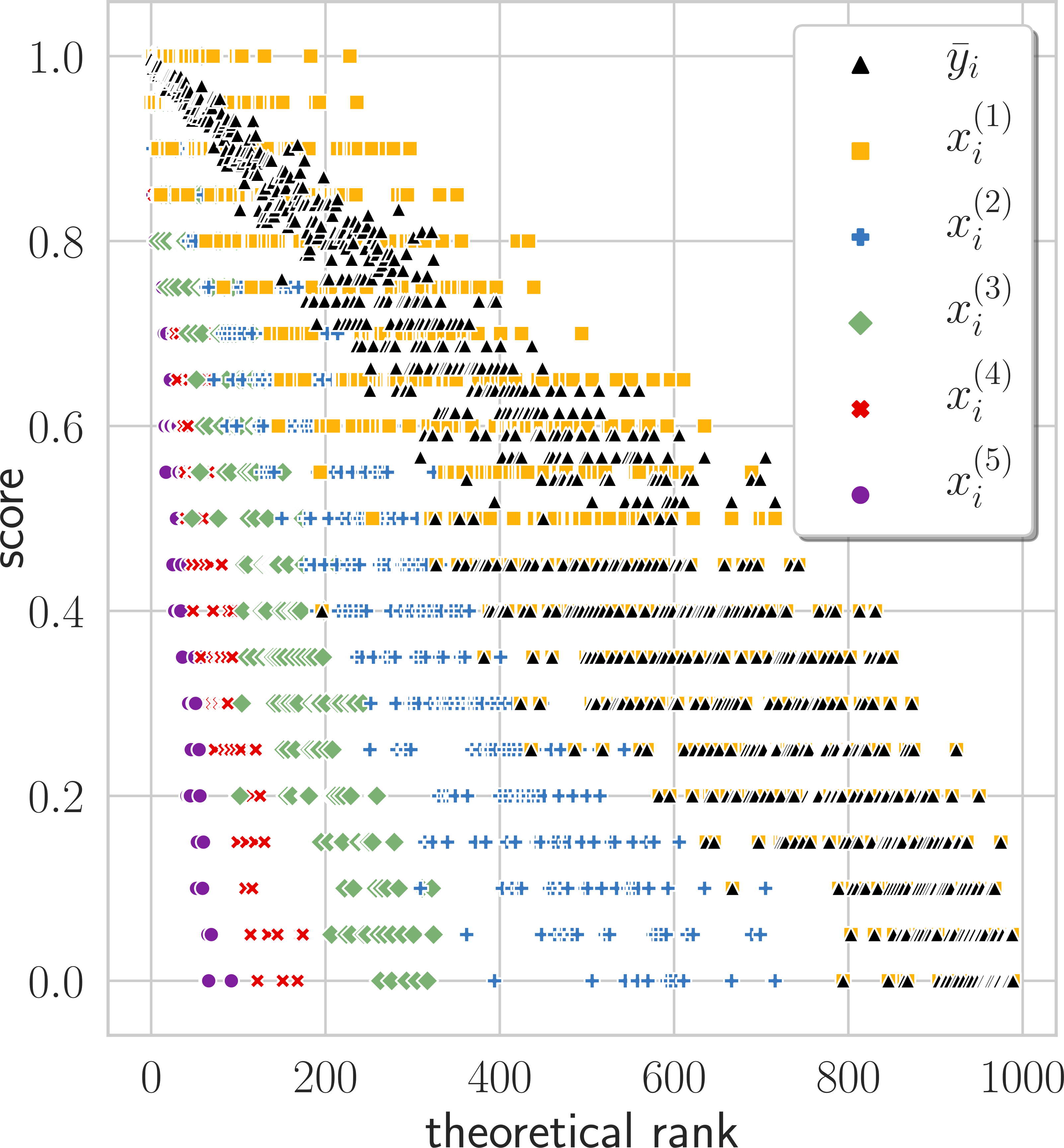}
\caption{The 
scores $x_i^{(k)}$ calculated in the first 5 ballots and the final approximation $\bar y_i$ are displayed 
as functions of the theoretical ranks. All values are obtained in the simulation described in the text.
} \label{fig:all_scores}
\end{figure}

Figure~\ref{fig:all_scores} displays the scores $x_i^{(k)}$ calculated in the first 5 ballots and the final approximation $\bar y_i$, obtained in a simulation based on the adaptive approach with exponential underlying similarity and the parameters described above;
the figure clearly shows that, as desired, the $\bar y_i$ precision is substantially larger for top ranks.

\section{Conclusion \& Future Work}

In this paper, we provided a protocol for the construction -- based on adaptive pairwise comparisons 
and tailored on the available resources -- of a dataset, which can be used to test or validate any relatedness-based domain-specific semantic model and which is optimized to be particularly accurate in top-rank evaluation.
Moreover, we defined the metrics $\rho_{\rm w}$ and $\tau_{\rm w}$, extensions of well-known ranking correlation coefficients, to evaluate a semantic model via the aforementioned dataset by taking into account the greater significance of top ranks.
Finally, we defined a stochastic transitivity 
model to simulate semantic-driven pairwise comparisons,
which allows tuning the parameters of the data collection approach and which confirmed a significant increase in the performance metrics $\rho_{\rm w}$ and $\tau_{\rm w}$ 
of the proposed  adaptive approach
compared with the uniform approach (see Table~\ref{tab:results}).
%
%

As 
future work, we plan to collect human annotations (\emph{i}) to 
test the proposed data collection approach on real data and (\emph{ii}) to 
assess the validity and estimate the parameters of the proposed stochastic transitivity model.
Additional future investigations may include a deeper analysis of the mathematical and statistical properties of the weighted coefficients $\rho_{\rm w}$, $\tau_{\rm w}$, as well as a rigorous derivation of the optimal values for the parameters of the data collection approach.


%

\bibliographystyle{acl_natbib}
\bibliography{SemanticEval}

\end{document}